\documentclass[journal]{IEEEtran} %lettersize
\usepackage{booktabs}
\usepackage{multirow}
\usepackage{comment}

\usepackage{cite}
\ifCLASSINFOpdf
    \usepackage[pdftex]{graphicx}
\else
\fi
\usepackage{amsmath}
\usepackage{algorithmic}
\usepackage{array}
\ifCLASSOPTIONcompsoc
  \usepackage[caption=false,font=normalsize,labelfont=sf,textfont=sf]{subfig}
\else
  \usepackage[caption=false,font=footnotesize]{subfig}
\fi
\usepackage{url}
\usepackage{xurl}
\usepackage{multirow}
\usepackage{hyperref}
\usepackage{enumitem}
\usepackage[linesnumbered,lined,ruled]{algorithm2e}
\usepackage[export]{adjustbox}
\usepackage{amsmath}
\usepackage{fontawesome5}
\usepackage{cleveref}
\usepackage{soul}
	
\usepackage{color, colortbl}

\usepackage{tikz,siunitx}
\usetikzlibrary{shapes.geometric,shapes.symbols}
\definecolor{cadmiumgreen}{rgb}{0.0, 0.42, 0.24}
\definecolor{mediumcandyapplered}{rgb}{0.89, 0.02, 0.17}
\definecolor{cottoncandy}{rgb}{1.0, 0.74, 0.85}
\definecolor{lightpink}{rgb}{1.0, 0.71, 0.76}
\definecolor{pastelpink}{rgb}{1.0, 0.82, 0.86}

\renewcommand\hl[1]{#1}  %removes \hl color

\begin{document}
\title{Towards Exploring Fairness in Visual Transformer based Natural and GAN Image Detection Systems}

\author{Manjary~P.~Gangan\IEEEauthorrefmark{1},~%\IEEEmembership{Member,~IEEE,}
        Anoop~Kadan\IEEEauthorrefmark{2},~%\IEEEmembership{Fellow,~OSA,}
        and~Lajish~V~L\IEEEauthorrefmark{1}%,~\IEEEmembership{Life~Fellow,~IEEE}% <-this % stops a space
%\thanks{This work was supported by the Women Scientist Scheme-A (WOS-A) for Research in Basic/Applied Science from the Department of Science and Technology (DST) of the Government of India under the Grant SR/WOS-A/PM-62/2018}
\\
\IEEEauthorblockA{\IEEEauthorrefmark{1}Department of Computer Science, University of Calicut, India}\\
\IEEEauthorblockA{\IEEEauthorrefmark{2}School of Psychology, University of Southampton, United Kingdom}
\thanks{Corresponding authors: Manjary~P.~Gangan and Anoop Kadan (e-mail: manjaryp\textunderscore dcs@uoc.ac.in, a.kadan@soton.ac.uk).}% <-this % stops a space
%\thanks{This work has been submitted to the IEEE for possible publication. Copyright may be transferred without notice, after which this version may no longer be accessible.}
}

\maketitle

\begin{abstract}

Image forensics research has recently witnessed a lot of advancements towards developing computational models capable of accurately detecting natural images captured by cameras and GAN generated images. However, it is also important to ensure whether these computational models are fair enough and \hl{do not} produce biased outcomes that could eventually harm certain societal groups or cause serious security threats. Exploring fairness in image forensic algorithms is an initial step towards mitigating these biases. This study \hl{explores} bias in visual transformer based image forensic algorithms that classify natural and GAN images, since visual transformers are recently being widely used in image classification based tasks, including in the area of image forensics. The proposed study procures bias evaluation corpora to analyze bias in gender, racial, affective, and intersectional domains using a wide set of individual and pairwise bias evaluation measures. Since the robustness of the algorithms against image compression is an important factor to be considered in forensic tasks, this study also analyzes the impact of image compression on model bias. Hence to study the impact of image compression on model bias, a two-phase evaluation setting is followed, where the experiments are carried out in uncompressed and compressed evaluation settings. The study could identify bias existences in the visual transformer based models distinguishing natural and GAN images, and also observes that image compression impacts model biases, predominantly amplifying the presence of biases in class GAN predictions.

\end{abstract}


%\ifCLASSOPTIONpeerreview
%\fi
%\IEEEpeerreviewmaketitle

\section{Introduction}
\label{sec_3bias_intro}


%\IEEEpubidadjcol


\IEEEPARstart{F}{airness} studies are recently gaining large interest in the research community since the machine learning based computational models are reported to have \hl{biases} in their outputs \cite{tian2022image}. These biases in the models can impact society by harming or denying opportunities to certain social groups of people \cite{ntoutsi2020bias}. Fairness studies report that these algorithmic biases can originate from training data, model representations, downstream tasks, etc., and accordingly, there are different kinds of algorithmic biases including data bias, model learning bias, downstream task level \hl{bias}, etc., \cite{anoop2022towards}. 

Easy availability of \hl{image-acquiring} devices, massive publicly accessible image datasets, rapid progress\hl{,} and a wide variety of generative algorithms and user-friendly easily available apps generating \hl{high-quality} super realistic images have drastically increased the production of GAN images all around. Beyond artistic and entertainment purposes, such GAN images are also seen to create some critical and harmful societal issues, such as evidence for supporting fake news, defamation, generating fake nude photographs, false light portrayals, \cite{anoop2019leveraging,karnouskos2020artificial,kietzmann2020deepfakes}, etc. Hence a lot of studies are reported proposing various methods for distinguishing GAN images from natural images captured by a camera, which can help to understand or even to serve as evidence to prove image authenticity \cite{afchar2018mesonet,li2020face,manjary2022distinguishing,gangan2023robust}. Although there is a lot of research in this area of distinguishing natural and GAN images, there are only a very few studies that explore algorithmic bias in such image forensics systems \cite{hazirbas2021towards,pu2022fairness}. Exploring bias in image forensics systems is very significant because unfair forensic systems can lead images of certain social groups to be more likely predicted as GAN images even if they are actually natural images. Unfair models may also lead images of certain social groups to be more likely predicted as natural images even if they are actually GAN generated images, creating security concerns. Therefore it is essential to test the fairness of image forensics systems.

Motivated by the transformer based networks that were initially designed dedicatedly for natural language based tasks, vision transformers were developed that handle images as sequences of patches \cite{dosovitskiy2021image}. Recently, visual transformer based models have drawn considerable attention due to their impressive performances for a variety of downstream tasks, for example, transformer based models for image classification (e.g. ViT \cite{dosovitskiy2021image}), object detection (e.g. DEtection TRansformer (DETR) \cite{carion2020end}, RT-DETR \cite{lv2023detrs}, YOLOS \cite{fang2021you}, ViT-YOLO \cite{zhang2021vit}), segmentation (e.g. SegFormer \cite{xie2021segformer}, Segmenter \cite{strudel2021segmenter}, image generation (e.g. Transgan \cite{jiang2021transgan}), etc., \cite{khan2022transformers,han2022survey}.
The area of image forensics also reports many works in the literature, utilizing these visual transformers \cite{wodajo2021deepfake,coccomini2022combining,wang2022m2tr,coccomini2022cross,gangan2023robust}. Due to the recent widespread use of visual transformers in image forensics, this study tries to explore bias, if any, in visual transformers for the forensic task of distinguishing natural (or real) and GAN generated images. 

\begin{figure*}[!htb]
\centering
\includegraphics[width=\linewidth]{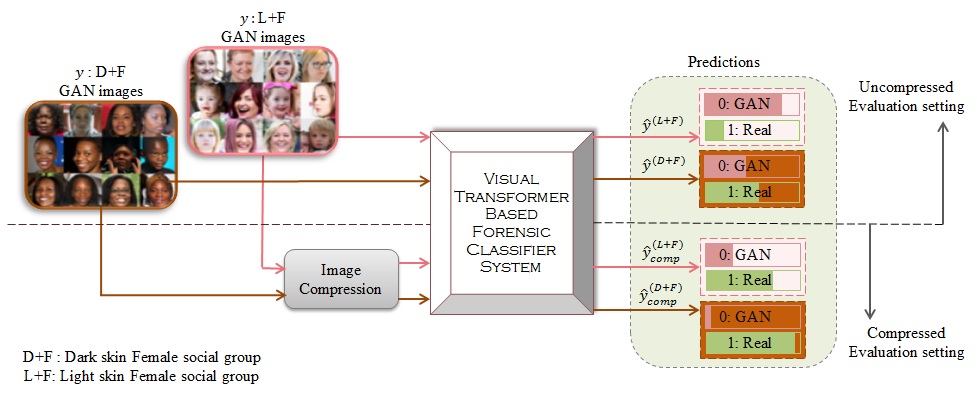}
\caption{The overall architecture of \hl{the} proposed work. The visual transformer based forensic classifier system is \hl{elaborated} in figure \ref{fig_3bias_model}}.
\label{fig_3bias_architecture}
\end{figure*}

Images shared through social media websites, unlike other post-processing operations, almost always go through compression knowingly or unknowingly \cite{chuman2017image}. Also, to deceive the forensic models detecting GAN images and to spread fake news, these images are usually compressed and propagated through social media \cite{wu2021towards,anoop2019leveraging}. Therefore, in the image forensic task of detecting natural and GAN generated images, the robustness of forensic algorithms towards post-processing operations, particularly image compression, is a very important factor to be considered. Hence, studies in the literature that build high performance GAN image detector systems also analyze the robustness of those models \cite{gangan2023robust}. Most studies report a high accuracy drop for the models, in compressed scenarios \cite{manjary2022distinguishing,gangan2023robust}. In this regard, one of the interests of this study, apart from identifying bias in visual transformers based classification of natural and GAN images, is to explore whether image compression impacts model bias. That is, this study focuses on two research objectives: \hl{(1) Do visual transformers produce algorithmic bias or unfairness in their predictions when utilized for the task of distinguishing natural and GAN generated images and, (2) does image compression impact or amplify algorithmic biases in these model predictions? For example, this study intends to explore whether any social groups (such as male or female, dark skin or light skin) are more likely predicted as natural images or as GAN generated images; and how image compression impacts the algorithmic biases (if any) towards these social groups.} To study these objectives, this work conducts bias analysis experiments in two evaluation settings, one in the original uncompressed evaluation setting and the other in the compressed setting, using the same set of evaluation measures. This helps to understand and identify any bias in the transformer based models and also to analyze whether model bias is impacted by image compression. Figure \ref{fig_3bias_architecture} shows the entire architecture of the proposed work, with an example set of input images and prediction scenarios to better understand the workflow and how this study conducts the bias exploration. This example only depicts the case of analyzing bias in GAN images\footnote{The GAN images in this example are collected from StyleGAN2 \cite{karras2020analyzing} generated images}, but the study considers analysis over both natural and GAN classes of images. 

The major contributions of the proposed work are:
\begin{itemize}
    \item \hl{This work explores algorithmic fairness in image forensic systems employing popular visual transformers, viz. ViT, CvT, and Swin for the task of distinguishing natural and GAN generated images.}
    \item \hl{The work analyzes algorithmic fairness in four different domains such as gender, racial, affective, and intersectional domains by procuring a bias evaluation corpora.}
    \item \hl{The work conducts extensive bias evaluation experiments using sets of individual and pairwise evaluation measures over the predictions of the forensic classifier systems, in each of the domains.}
    \item \hl{The work also tries to understand the impact of image compression on biases in forensic classifier systems by analyzing and comparing the existence of model biases, across uncompressed and compressed evaluation settings.}
\end{itemize}

The rest of the paper is organized as section \ref{sec_3bias_review} presents a brief survey on the works in literature that specifically analyze bias in image forensic tasks classifying natural and GAN generated images and explains the differences of the proposed study in the context of the works in the literature. Section \ref{sec_3bias_classification} discusses in detail the construction of transformer based models for the task of classifying natural and GAN images. Section \ref{sec_3bias_biasanalysis} explains in detail the evaluation domains, evaluation corpora, and evaluation measures used for bias analysis experiments. Section \ref{sec_3bias_results} \hl{presents} the results and discussions of both the uncompressed and compressed evaluation settings and finally section \ref{sec_3bias_conclu} presents the conclusions and future directions of the work.

\section{Related work and Our work in context}
\label{sec_3bias_review}

The boom of big data and expeditious progress in deep learning based models helped to bring up solutions to many fields of computing \cite{zhao2021multi,wang2020research,chen2020citywide,wang2020big,zhang2020blockchain,tey2022generative}. However, these data-greedy models are found to produce biased outcomes \cite{tian2022image,ntoutsi2020bias}. Many works are seen to be reported in the literature studying fairness in image based research problems, such as in the areas of face recognition \cite{buolamwini2018gender}, image classification \cite{schaaf2021towards}, medical image processing \cite{baxter2022bias}, etc. But comparably only a very few studies explore bias in forensics systems, and amongst those studies, most of them work on videos, i.e., Deep Fake videos. 
Trinh and Liu \cite{trinh2021examination} explore bias in three deep fake detection models Xception \cite{chollet2017xception}, MesoInception-4 \cite{afchar2018mesonet} and Face X-Ray \cite{li2020face}, using gender and race balanced Deep Fake face datasets. Their study observes high racial bias in the predictions of these Deep Fake detection models. They could also observe that one of the most popularly used datasets for training the models for Deep Fake detection, FaceForensics++ \cite{rossler2019faceforensics}, is also highly biased towards female Caucasian social groups.
Hazirbas et al. \cite{hazirbas2021towards} proposes a video dataset to analyze the robustness of top-winning five models of DFDC dataset \cite{dolhansky2020deepfake} for the domains gender, skin type, age, and lighting. They could observe that all the models are biased against \hl{dark-skinned} people and hence find that these five models are not generalizable to all groups of people.
Pu et al. \cite{pu2022fairness} explores gender bias in one of the Deep Fake detection models MesoInception-4, in the presence of certain make-up anomalies, using the FaceForensics dataset. Their study is centered on analyzing these models at various prominence levels of the anomaly in the female and male social groups. Their observations are that the model is biased towards both genders, but mostly towards the female group.
Xu et al. \cite{xu2023comprehensive} explores bias in three Deep Fake detection models EfficientNetB0 \cite{tan2019efficientnet}, Xception \cite{chollet2017xception}, and Capsule-Forensics-v2 \cite{nguyen2019use}, by conducting evaluations on five Deep Fake datasets which are annotated with 47 attributes including non-demographic and demographic attributes. Their observations state that these models are highly unfair towards many of these attributes.

\subsection{Proposed work in \hl{the} context of the literature}
\label{sec_3bias_proposedwork}

In the context of the previous works in the literature that analyze bias in image forensic algorithms classifying natural and GAN \hl{generated} images \cite{trinh2021examination,hazirbas2021towards,pu2022fairness,xu2023comprehensive}, the proposed work is the first work, to the best knowledge, that explores bias in transformer based image forensic models classifying natural and GAN generated images. Also, the proposed work is the first work, to the best knowledge, to study the role/impact of image compression in model biases. The work tries to unveil any existence of bias in gender, racial, affective, and even intersectional domains using a vast set of individual and pairwise evaluation measures, and sets aside the mitigation of these biases outside the scope of this work, for future studies.

\section{Classification of natural and GAN generated images}
\label{sec_3bias_classification}

This section discusses the dataset used to fine-tune the transformer based models and construction of \hl{the} visual transformer based forensic classifier system for the task of classifying natural and GAN generated images, \hl{which} are investigated for fairness in this study.

\subsection{Fine-tuning corpora}
\label{sec_3bias_finetuning}

To build transformer based forensic classifier systems that classify GAN and Real images, each of the pre-trained transformer based models are fine-tuned using a GAN versus Real image dataset that consists of a total of 10,000 images; each class containing 5000 images. The GAN images are collected from the StyleGAN2 image generative algorithm \cite{karras2020analyzing} and the Real class of images are collected from the Flickr-Faces-HQ (FFHQ) dataset \cite{karras2019style}; these datasets are popularly used in many related studies in the literature \cite{barni2020cnn,guo2022robust,manjary2022distinguishing,gangan2023robust}. Since works exploring algorithmic biases focus on performance differences between groups within domains in which bias study is conducted \cite{hazirbas2021towards,pu2022fairness,kadan2023blacks}, rather than focusing on \hl{the} performance of the models, in this study the total fine-tuning corpora is split in the ratio 6:2:2 for training, validation and testing, respectively, by referring the related works \cite{manjary2022distinguishing,gangan2023robust}.

\subsection{{Visual transformer based forensic classifier system}}
\label{sec_3bias_transformer}

This work tries to identify bias in three popular transformer based deep learning models, viz. Vision Transformer (ViT) \cite{dosovitskiy2021image}, Convolutional Vision Transformer (CvT) \cite{wu2021cvt} and Swin transformer \cite{liu2021swin}, for the task of classifying natural and GAN images. The ViT architecture divides the images into \hl{fixed-size} patches in order. These \hl{non-overlapping} patches are then linearly embedded. These embeddings along with the position embeddings of the patches and a learnable classification token are supplied to the transformer encoder block for classification task \cite{dosovitskiy2021image}. CvT architecture utilizes convolutions within the ViT architecture with an aim to improve the performance of ViT. The major difference includes using a set of transformers with convolutional token embedding, convolutional projection and convolutional transformer block \cite{wu2021cvt}. Swin transformer follows hierarchical architecture based on Shifted WINdow approach \cite{liu2021swin}. 

\begin{figure}[!htb]
\centering
\includegraphics[width=\linewidth]{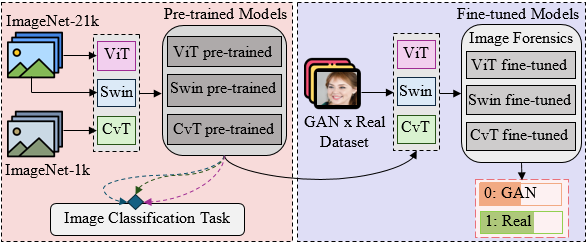}
\caption{{Visual transformer based forensic classifier system}}
\label{fig_3bias_model}
\end{figure}

The natural image versus GAN image classification is formulated as a \hl{two-class} classification task that can classify images under evaluation into either of the two classes GAN or Real. The classifiers are fed with the training data $x_1, x_2, \ldots, x_N$ ($x_i$ indicates i\textsuperscript{th} image in train data) and associated ground truth classes $y_1, y_2, \ldots, y_N$ ($y \in \{\text{GAN, Real}\}$) such that to find a best fitting model $M: y=M(x)$. To build the classifier models, the three pre-trained transformer networks {ViT, CvT\hl{,} and Swin,} are fine-tuned using the \hl{task-specific} GAN versus Real image dataset. A diagrammatic representation of the visual transformer based natural image versus GAN image classifier model is shown in figure \ref{fig_3bias_model}.

\subsection{\hl{Fine-tuning experimental settings}}

\hl{Fine-tuning experiments of the transformers are conducted on the deep learning workstation equipped with Intel Xeon Silver 4208 CPU at 2.10 GHz, 256 GB RAM, and two GPUs of NVIDIA Quadro RTX 5000 (16GB each), using the libraries Torch (version 1.13.1+cu116), PyTorch Lightning (version 1.9.0), Transformer (version 4.17.0), Tensorflow (version 2.8.0), and Keras (version 2.8.0). Table \mbox{\ref{table_3bias_parameters}} shows the model parameters.}

\begin{table}[!h]
\setlength{\tabcolsep}{3pt}
\centering
\caption{Model parameters}
\label{table_3bias_parameters}
%\rowcolors{1}{yellow}{yellow} % for highlighting the table 
\begin{tabular}{@{}lccc@{}}
\toprule
Parameter & ViT & CvT & Swin \\ \midrule
Network & ViT-Large & CvT-21 & Swin-Large \\
Patch size & 16 & - & 4 (window size: 7) \\
%Window size & - & - & 7 \\
\begin{tabular}[c]{@{}l@{}}Pre-training\\dataset\end{tabular} & ImageNet-21K\cite{ridnik2021imagenet} & ImageNet-1k\cite{russakovsky2015imagenet} & ImageNet-21K\cite{ridnik2021imagenet} \\
Input image & 224 $\times$ 224 & 224 $\times$ 224 & 224 $\times$ 224 \\
Learning rate & $2e-5$ & $2e-5$ & $2e-5$ \\
Batch size & 4 & 4 & 4 \\
Optimizer & Adam & Adam & Adam \\
Epoch & 25 & 25 & 25 \\
\begin{tabular}[c]{@{}l@{}}Trainable\\parameters\end{tabular} 
    & 303 M & 31.2 M & 194 M \\
\bottomrule
\end{tabular}
\end{table}
%
\begin{comment}
Table \ref{table_3bias_train_val_test_acc} shows the test accuracy of the fine-tuned transformer based models in classifying natural and GAN images. 
\begin{table}[!htb]
\centering
\caption{Fine-tuned model accuracies}
\label{table_3bias_train_val_test_acc}
\begin{tabular}{lccc}
\toprule
\multicolumn{1}{c}{Model} & \multicolumn{1}{c}{Total test accuracy} & \multicolumn{1}{c}{GAN accuracy} & \multicolumn{1}{c}{Real accuracy}  \\
\midrule
ViT & 91.75 & 94.9 & 88.6 \\
CvT & 99.60 & 99.6 & 99.6 \\
Swin & 99.70 & 99.9 & 99.5 \\
\bottomrule
\end{tabular}
\end{table}
%
\end{comment}

\section{Bias Analysis in forensic classifier systems}
\label{sec_3bias_biasanalysis}

This study tries to identify bias (if any), in the transformer based \textit{Natural image versus GAN generated image} classifier systems. Fairness analysis is conducted in the gender, racial, affective, and also in intersectional domains. Gender domain based bias analysis considers the female and male social groups, the racial domain considers the dark skin people and light skin people social groups, the affective domain considers the smiling face and \hl{non-smiling} face groups and intersectional bias analysis considers two domains simultaneously, such as dark skin female, light skin male, etc. Apart from analyzing bias by comparing the performances of each social group against the other using individual evaluation measures, this study also performs pairwise analysis of social groups. Bias analysis in this forensic task of classifying natural and GAN generated images using transformer based models is conducted using two categories of evaluation corpora, one consisting of the original uncompressed GAN and Real evaluation corpora and the other is the JPEG compressed version of the same evaluation corpora. That is, in the first phase of bias analysis, the transformer based models are evaluated over the uncompressed evaluation corpora using a set of evaluation measures, and in the second phase of analysis the same evaluation corpora is JPEG compressed with a quality factor of 90 and analyzed using the same evaluation measures. The details of evaluation corpora and evaluation measures are detailed below.

\subsection{Evaluation {domains and evaluation corpora}}
\label{sec_3bias_evaluationdomain}

This work procures an evaluation corpora for bias analysis with respect to gender, racial, and affective domains. To procure the evaluation corpora we utilize Natural images from the FFHQ dataset \cite{karras2019style} and GAN images from the StyleGAN2 \cite{karras2020analyzing} generated images. From both Natural and GAN generated images we collected 1000 female face images and 1000 male face images each for the gender bias analysis, 1000 dark skin and 1000 light skin face images for racial bias analysis, and 1000 smiling and 1000 non smiling face images for affective bias analysis. This also gives chances for intersectional bias analysis with 500 images each in the category of dark skin female, dark skin male, light skin female, and light skin male faces. %A sample set of GAN images from the evaluation corpora used in this study is provided in figure \ref{fig_3bias_eval_corpora} (even though the real class of images in the evaluation corpora are collected from the publicly available FFHQ dataset which is properly cited as \cite{karras2019style}, we avoid portraying the images of real people for showing the examples of each social groups, and only use the sample images from the class GAN). 

%
\begin{comment}
%
\begin{figure*}[!h]
\centering
\subfloat[Dark skin female]
    {\includegraphics[width=.15\textwidth,keepaspectratio]{figures_3bias/073142.png}\label{fig_3bias_eval_corpora_df}}
%\hfil
\subfloat[Light skin female]
    {\includegraphics[width=.15\textwidth,keepaspectratio]{figures_3bias/098460.png}\label{fig_3bias_eval_corpora_lf}}
%\hfil
\subfloat[Dark skin male]
    {\includegraphics[width=.15\textwidth,keepaspectratio]{figures_3bias/065453.png}\label{fig_3bias_eval_corpora_dm}}
%\hfil
\subfloat[Light skin male]
    {\includegraphics[width=.15\textwidth,keepaspectratio]{figures_3bias/097042.png}\label{fig_3bias_eval_corpora_lm}}
%\hfil
\subfloat[Non smiling face]
    {\includegraphics[width=.15\textwidth,keepaspectratio]{figures_3bias/095551.png}\label{fig_3bias_eval_corpora_ns}}
%\hfil
\subfloat[Smiling face]
    {\includegraphics[width=.15\textwidth,keepaspectratio]{figures_3bias/097014.png}\label{fig_3bias_eval_corpora_s}}
\caption{A sample of GAN face images from the evaluation corpora used in this study}
\label{fig_3bias_eval_corpora}
\end{figure*}
%
\end{comment}
%

\subsection{Evaluation {measures}}
\label{sec_3bias_evaluationmeasures}

Bias analysis in this study focuses on comparing the classification performance of the transformer based models over different social groups (or groups) within the same domain using certain evaluation measures. These analyses are performed to compare social groups within a single domain (e.g. Male vs. Female in \hl{the} gender domain) as well as to compare social groups within intersectional domains (e.g. Dark skin Male vs. Light skin Male). \hl{Apart from the measures that evaluate individual social groups such as total accuracy, GAN and real class accuracies, false positive rate, and false negative rate, this study also utilizes pairwise evaluation measures such as average confidence score, demographic parity, and equal opportunity, to quantify bias associated with a pair of social groups in single domain or intersectional groups in a domain. The measures considering individual social groups in a domain and pairwise measures considering two social groups simultaneously are detailed below.}

\subsubsection{Individual measures considering a single group within a domain}

These measures are defined by the probability of correct and incorrect classifications in a social group within a domain. Social groups over which the individual measures are evaluated include, Female (F) and Male (M) social groups in the gender domain, Dark skin (D) and Light skin (L) social groups in the racial domain, Non-smiling (Ns) and Smiling (S) groups in the affective domain, and Dark skin Female (DF), Dark skin Male (DM), Light skin Female (LF) and Light skin Male (LM) groups in the intersectional domain.

\begin{itemize}[leftmargin=5mm]
    \item \hl{Total accuracy} \cite{buolamwini2018gender,dressel2018accuracy}: This popular classification measure computes the total classification accuracy of a model over a social group in a domain. Total accuracy gives the percentage of images in a social group that is correctly classified into the natural image category and the GAN generated image category. 
    \begin{align}
        Acc = \frac{TP + TN}{ TP + TN + FP + FN}
    \end{align}
    where, TP and TN are the number of true positives and true negatives, and FP and FN denote false positives and false negatives, respectively.
    \item \hl{GAN accuracy}: This measure gives the accuracy of the class GAN images, i.e., the number of GAN images correctly classified as GAN images. This measure gives the True Positive Rate (TPR) \cite{dressel2018accuracy} of the model
    \begin{align}
        Acc_{gan} = \frac{TP}{TP + FN}
    \end{align}
    \item \hl{Real accuracy}: This measure gives the accuracy of the class of natural images, i.e., the number of natural images correctly classified as natural images. This measure is the True Negative Rate (TNR) \cite{dressel2018accuracy} of the model.
    \begin{align}
        Acc_{real} = \frac{TN}{TN + FP}  
    \end{align}
    \item \hl{False Positive Rate (FPR)} \cite{buolamwini2018gender,dressel2018accuracy}: For this classification task, FPR gives the ratio of Real images misclassified as GAN images, among the total number of Real images.
    \begin{align}
        FPR = \frac{FP}{FP +TN} 
    \end{align}
    \item \hl{False Negative Rate (FNR)} \cite{dressel2018accuracy}: FNR gives the ratio of GAN images misclassified as Real images, among the total number of GAN images.
    \begin{align}
        FNR = \frac{FN}{TP +FN} 
    \end{align}    
\end{itemize}
During evaluation, the results obtained for each of these individual measures across the social groups within a domain are correspondingly compared, rather than looking for ideal high classification results.

\subsubsection{Pairwise evaluation measures \hl{considering} a pair of groups within a domain}

The pairwise evaluations are computed on a pair of social groups $g^{(a)}$ and $g^{(b)}$ within a domain. $y(g^{(a)}_i)$ indicates the ground truth class of i\textsuperscript{th} image in the social group $g^{(a)}$ (for $i \in A$), and $y(g^{(b)}_j)$ indicates the ground truth class of j\textsuperscript{th} image in the group $g^{(b)}$ (for $j \in B$), where $A$ and $B$ indicates total number of instances in the social groups $g^{(a)}$ and $g^{(b)}$, respectively. Also, $y_{class}(g^{(a)}_i)$ and $y_{class}(g^{(b)}_j)$ indicate the corresponding prediction classes, and $y_{score}(g^{(a)}_i)$ and $y_{score}(g^{(b)}_j)$ indicate prediction intensities (confidence scores of prediction), of $g^{(a)}$ and $g^{(b)}$, respectively. Pairwise measures are evaluated over the pairs, Female vs. Male (F$\times$M) in gender domain, Dark skin vs. Light skin (D$\times$L) in racial domain, Non-smiling vs. Smiling (Ns$\times$S) in affective domain, and Dark Female vs. Dark Male (D+F $\times$ D+M), Light Female vs. Light Male (L+F $\times$ L+M), Dark Female vs. Light Female (D+F $\times$ L+F), Dark Male vs. Light Male (D+M $\times$ L+M), Dark Female vs. Light Male (D+F $\times$ L+M)  and Light Female vs. Dark Male (L+F $\times$ D+M) in the intersectional domain.

\begin{itemize}[leftmargin=5mm]
    \item \hl{Average Confidence Score (ACS)} \cite{kadan2023blacks}: This measure is computed using the ratio between average prediction intensities of the two social groups under evaluation.
    \begin{align}
        ACS = 1- \frac{\frac{1}{A}\left(\sum_{i=1}^A y_{score}(g^{(a)}_i) \right)}{\frac{1}{B}\left(\sum_{i=1}^B y_{score}(g^{(b)}_j) \right)} 
    \end{align}  
    An ideal unbiased scenario gives ACS = 0 for a pair. Positive values of ACS show that the prediction intensities of the social group $g^{(a)}$ are lower than $g^{(b)}$, whereas negative ACS indicates that the prediction intensities of the social group $g^{(a)}$ are higher than $g^{(b)}$.
    \item \hl{Demographic Parity (DP)} \cite{kadan2023blacks, tian2022image}: This is one of the popular measures to quantify bias in a classification model, by analyzing similarity (or dissimilarity) in the classifications of the model for two social groups in a domain.
    \begin{align}
        DP = \frac{P\left( y_{class}(g^{(a)}_i) = c \;|\; z = g^{(a)} \right)}{P\left( y_{class}(g^{(b)}_j) = c \;|\; z = g^{(b)} \right)}
    \end{align} 
    where, $P\left( y_{class}(g^{(a)}_i) = c \;|\; z = g^{(a)} \right)$ and $P\left( y_{class}(g^{(b)}_j) = c \;|\; z = g^{(b)} \right)$ are the probabilities of the groups $g^{(a)}$ and $g^{(b)}$, respectively, for being classified into a class $c \in \text{(GAN, Real)}$ where, in the $g^{(a)}$ $\times$ $g^{(b)}$ pair, $g^{(a)}$ is the group with higher probability. That is, the measure DP recommends that the probability of predicting a class $c$ needs to be similar for both the social groups $g^{(a)}$ and $g^{(b)}$ within a domain. Hence, an ideal unbiased case is indicated by DP = 1 for a pair, and lower values of DP indicate higher bias. A threshold of  0.80 is commonly used for identifying lower DP values, indicating high model bias \cite{hardt2016equality}.
    \item \hl{Equal Opportunity (EO)} \cite{tian2022image,hardt2016equality}: This measure is also similar to DP, but EO considers the ground truth in addition to the predicted classes. 
    \begin{align}
    DP = \frac{P\left( y_{class}(g^{(a)}_i) = c \; \&\& \; y(g^{(a)})=c \;  \;|\; z = g^{(a)} \right)}{P\left( y_{class}(g^{(b)}_j) = c \; \&\& \; y(g^{(b)})=c \;|\; z = g^{(b)} \right)}
    \end{align}
    where, $y(g^{(a)}=c)$ and $y(g^{(b)}=c)$ indicates the ground truth class $c$ of group $y(g^{(a)})$ and $y(g^{(b)})$. Similar to DP, an ideal unbiased case is indicated by EO = 1 for a pair, and lower values of EO indicate higher bias. 
\end{itemize}

\section{Results and analysis}
\label{sec_3bias_results}

Since this study follows a two-phase evaluation setting, where the bias evaluation experiments are carried out in both the uncompressed and compressed settings, this section initially analyses the results of bias evaluation of each of the transformer based models, ViT, CvT, and Swin over the original uncompressed evaluation corpora and later the results of bias evaluation of the models over the compressed evaluation corpora.

\subsection{Bais analysis in the \textbf{uncompressed} evaluation setting}
\label{sec_3bias_results_uncompressed}

\subsubsection{\textbf{ViT}}

Bias evaluation results of the ViT based model in the uncompressed setting \hl{are} shown in table \ref{table_3bias_vit}. The top portion of the table presents the results of individual measures of bias analysis of ViT and the bottom portion presents the results of pairwise measures of bias analysis, within gender, race, affective, and intersectional domains.

\begin{table}[!h]
\centering
\caption{Evaluation results of \textbf{ViT} in \textbf{uncompressed} setting}
\label{table_3bias_vit}
\resizebox{\linewidth}{!}{
\setlength{\tabcolsep}{3pt}
\begin{tabular}{@{}c|cc|cc|cc|cccc@{}}
\toprule
\multicolumn{10}{c}{\textbf{Individual measures based analysis}} \\
\midrule
\multicolumn{1}{c|}{\multirow{2}{*}{\hl{Measure}}} &
  \multicolumn{2}{c|}{Gender} &
  \multicolumn{2}{c|}{Race} &
  \multicolumn{2}{c|}{Affective} &
  \multicolumn{4}{c}{Intersection} \\ \cmidrule(l){2-11} 
\multicolumn{1}{c|}{} &
  \multicolumn{1}{c|}{F} &
  \multicolumn{1}{c|}{M} &
  \multicolumn{1}{c|}{D} &
  \multicolumn{1}{c|}{L} &
  \multicolumn{1}{c|}{Ns} &
  \multicolumn{1}{c|}{S} &
  \multicolumn{1}{c|}{D+F} &
  \multicolumn{1}{c|}{D+M} &
  \multicolumn{1}{c|}{L+F} &
  \multicolumn{1}{c}{L+M} \\ \midrule
Acc & 88.20  & 92.65 & 92.50  & 88.35 & 92.70  & 89.75 & 90.70  & 94.30  & 85.70  & 91.00  \\ \midrule
Acc\textsubscript{gan} & 93.10  & 92.70  & 91.50  & 94.30  & 93.90  & 95.10  & 89.80  & 93.20  & 96.40  & 92.20  \\ \midrule
Acc\textsubscript{real} & 83.30  & 92.60  & 93.50  & 82.40  & 91.50  & 84.40  & 91.60  & 95.40  & 75.00 & 89.80  \\ \midrule
FPR & 0.167 & 0.074 & 0.065 & 0.176 & 0.085 & 0.156 & 0.084 & 0.046 & 0.250  & 0.102 \\ \midrule
FNR  & 0.069 & 0.073 & 0.085 & 0.057 & 0.061 & 0.049 & 0.102 & 0.068 & 0.036 & 0.078 \vspace{5pt}\\ %\bottomrule
\end{tabular}}
\resizebox{\linewidth}{!}{
\setlength{\tabcolsep}{1.5pt}
\begin{tabular}{@{}c|c|c|c|cccccc@{}}
\toprule
\multicolumn{10}{c}{\textbf{Pairwise measures based analysis}} \\
\midrule
\multicolumn{1}{c|}{} & \multicolumn{1}{c|}{Gender} & \multicolumn{1}{c|}{Race} & \multicolumn{1}{c|}{Affect} & \multicolumn{6}{c}{Intersection} \\ \midrule
\multicolumn{1}{c|}{} &
  \multicolumn{1}{c|}{F$\times$M} &
  \multicolumn{1}{c|}{D$\times$L} &
  \multicolumn{1}{c|}{Ns$\times$S} &
  \multicolumn{1}{c|}{\begin{tabular}[c]{@{}l@{}} D+F $\times$ \\ D+M \end{tabular}} &
  \multicolumn{1}{c|}{\begin{tabular}[c]{@{}l@{}} L+F $\times$ \\ L+M \end{tabular}} &
  \multicolumn{1}{c|}{\begin{tabular}[c]{@{}l@{}} D+F $\times$ \\ L+F \end{tabular}} &
  \multicolumn{1}{c|}{\begin{tabular}[c]{@{}l@{}} D+M $\times$ \\ L+M \end{tabular}} &
  \multicolumn{1}{c|}{\begin{tabular}[c]{@{}l@{}} D+F $\times$ \\ L+M \end{tabular}} &
  \multicolumn{1}{c}{\begin{tabular}[c]{@{}l@{}} L+F $\times$ \\ D+M \end{tabular}}  \\ \midrule
\multicolumn{10}{c}{GAN} \\
\midrule
\hl{ACS} & -0.0036 & +0.0173 & +0.0049 & +0.0173 & -0.0232 & +0.0368 & -0.0029 & +0.0145 & -0.0202 \\ \midrule
\hl{DP}  & 0.9117  & 0.8758  & 0.9250  & 0.9959  & 0.8435  & \textbf{0.7989}  & 0.9551  & 0.9590  & \textbf{0.7956}  \\ \midrule
\hl{EO} & 0.9957  & 0.9703  & 0.9874  & 0.9635  & 0.9564  & 0.9315  & 0.9893  & 0.9740  & 0.9668  \\ \midrule
\multicolumn{10}{c}{Real} \\ \midrule
ACS & +0.0252 & -0.0212 & -0.0200 & +0.0139 & +0.0398 & -0.0368 & -0.0095 & +0.0045 & +0.0489 \\ \midrule
DP  & 0.9029  & 0.8637  & 0.9150  & 0.9961  & \textbf{0.8053}  & \textbf{0.7721}  & 0.9550  & 0.9588  & \textbf{0.7691}  \\ \midrule
EO  & 0.8996  & 0.8813  & 0.9224  & 0.9602 & 0.8352  & 0.8188  & 0.9413  & 0.9804  & 0.7862  \\ \bottomrule
\end{tabular}}
\end{table}

While looking into the results of individual measures (in the top portion of table \ref{table_3bias_vit}), in the gender domain, the total model accuracy (Acc) of ViT over the female group (F) is less than the male group (M) by 4.45 percentage points, indicating biased prediction. This bias is observed to be very high for class Real (Acc\textsubscript{real}), i.e. accuracy of the female group is less than male by 9.3 percentage points, indicating high gender bias against the female social group. Whereas, for class GAN (Acc\textsubscript{gan}), the accuracy of the male group is less than female only by a very small value of 0.4 percentage points, a negligible difference to indicate any bias. Also, in the gender domain, the measure FPR is higher for the female group than \hl{the} male. This indicates Real images of females are more likely to be misclassified as GAN generated images than those of males (an observation similar to the one reported in \cite{trinh2021examination}). Whereas, the very low difference in FNR values between male and female groups indicates negligible chances that GAN images of males get misclassified as Real images.

In the racial domain, the total accuracy of ViT over the light skin (L) group is less than dark skin (D) by 4.15 percentage points, indicating biased prediction against light skin people, which is much more evident in the case of class Real {with} a difference of 11.1 percentage points, indicating high racial bias against light skin people. Whereas, in the case of {class} GAN, the accuracy of the dark skin group is less than light skin by 2.8 percentage points, indicating bias against dark skin. The measure FPR shows a higher value for light skin group than dark skin, which indicates Real images of light skin people are more likely to be misclassified as GAN images, and FNR indicates slight chances for GAN images of dark skin people being misclassified as Real images.

In the affective domain, the total accuracy of ViT over the group of smiling faces (S) is less than non-smiling faces (Ns) by 2.95 percentage points. A similar pattern is shown in class Real, with a difference of 7.1 percentage points, indicating affective bias against the group with smiling faces. Whereas in the case of class GAN, the accuracy of smiling faces is higher than non-smiling faces by 1.2 percentage points. FPR shows high value for smiling faces, which indicates Real images of smiling people are {more} likely to be misclassified as GAN images. The slightly higher values of FNR for non-smiling faces indicate slight chances for GAN images of non-smiling faces being misclassified as Real images.

In the intersectional domain, it can be observed that the total accuracy varies across different intersectional groups. {The highest total} accuracy is observed for dark skin male group {(D+M)}, and lowest for light skin female {(L+F)}, {with} a difference of 8.6 percentage points, {which indicates bias against light skin female group}. Whereas for class GAN, {an} accuracy of 96.4 {percent} is obtained for light skin female group, which is the highest accuracy obtained across various groups {among both classes and even compared to the total accuracy}. The lowest accuracy in class GAN is for the dark skin female group {(D+F)}, a difference of 6.6 percentage points compared to the highest accuracy group, indicating biased prediction. In class Real, the highest accuracy is obtained for dark skin male group and the lowest accuracy of 75.0 percent is obtained for light skin female group, which is the lowest accuracy obtained across various groups among both classes and even compared to the total accuracy. That is, both these groups have a very high difference of 20.4 percentage points, indicating very large intersectional bias against light skin female \hl{group}. FPR stands highest for the light skin female group indicating \textit{Real images of light skin females have a very high probability of being misclassified as GAN images}. FNR is highest for the dark skin female group indicating \textit{GAN images of dark skin females have a very high probability of being misclassified as Real images}.

The bottom portion of the same table \ref{table_3bias_vit} presents the results of pairwise measures of bias analysis of ViT for both GAN and Real classes. In the gender domain, for class GAN, the negative value of the measure ACS for the Female vs. Male pair (F$\times$M) shows that the prediction intensities of the female group are higher than males. The measure DP has a low value, but since it is not less than the threshold of 0.80 this measure does not report bias in the Female vs. Male pair. The measure EO has a high value and does not report gender bias in class GAN predictions. For \hl{the} class Real, positive ACS for Female vs. Male pair shows that the prediction intensities of the male group are higher than females. The measures DP and EO have low values. But since DP is not lower than the threshold \hl{of} 0.80, it does not report gender bias in class Real predictions.

In the racial domain, for class GAN, the positive ACS value for Dark skin vs. Light skin pair {(D$\times$L)} shows that the prediction intensities of the light skin group are higher than dark skin. The measure DP has a low value, but since it is not less than the threshold of 0.80\hl{,} this measure does not report racial bias. EO has a high value and does not report racial bias in the class GAN predictions. For the class Real, negative ACS for the pair shows that the prediction intensities of the dark skin group are higher than light skin. The measures DP and EO have low values, where DP is not lower than the threshold of 0.80\hl{,} and hence do not report racial bias in the class Real predictions.

In the affective domain, for class GAN, the positive ACS value for the Non-smiling vs. Smiling pair {(Ns$\times$S)} shows that the prediction intensities of the smiling group are higher than the non-smiling group. The measure DP has a low value, but since it is not less than the threshold of 0.80\hl{,} this measure does not report bias in this pair. EO has a high value and does not report bias in {the class} GAN predictions of this pair. For the class Real, negative ACS for the pair shows that the prediction intensities of the non-smiling group are higher than the smiling group. The measures DP and EO have low values. But since DP is not lower than the threshold of 0.80, {this measure} do not report bias in Real predictions of this pair.

\begin{figure*}[!htb]
\centering
    \includegraphics[width=.7\textwidth,keepaspectratio]{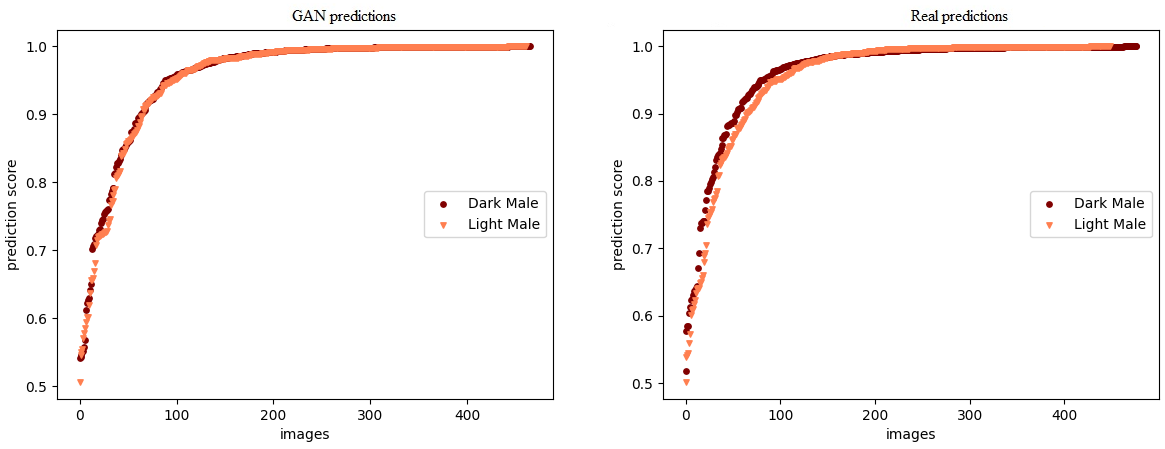}\label{fig_3bias_result_vit_dmxlm}
\caption{{Prediction intensity plots of an \textbf{unbiased} intersectional pair Dark skin Male vs. Light skin Male (D+M $\times$ L+M) in the bias evaluation corpora}} 
\label{fig_3bias_result_vit_unbiased}
\end{figure*}

\begin{figure*}[!htb]
\centering
    \includegraphics[width=.7\textwidth,keepaspectratio]{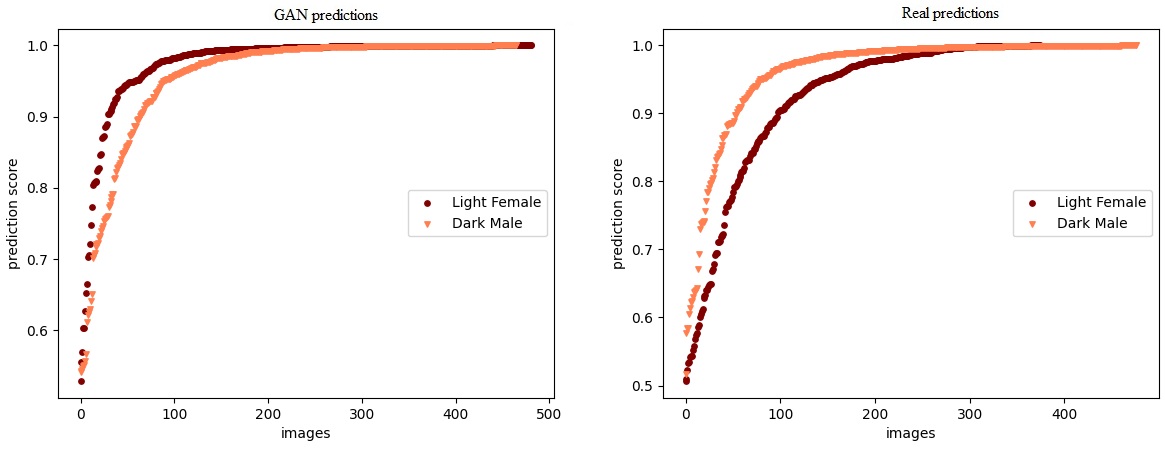}\label{fig_3bias_result_vit_lfxdm}
\caption{{Prediction intensity plots of a \textbf{biased} intersectional pair Light skin Female vs. Dark skin Male (L+F $\times$ D+M) in the bias evaluation corpora}} 
\label{fig_3bias_result_vit_biased}
\end{figure*}

In the intersectional domain, for the class GAN, the measure DP is very low for the pairs involving light skin female {group}, i.e., \{Light skin Female vs. Light skin Male\} (\underline{L+F} $\times$ L+M), \{Dark skin Female vs. Light skin Female\} (D+F $\times$ \underline{L+F}) and \{Light skin Female vs. Dark skin Male\} (\underline{L+F} $\times$ D+M). Particularly for the pairs \{Dark skin Female vs. Light skin Female\} and \{Light skin Female vs. Dark skin Male\}, {the measure} DP is less than the threshold of 0.80, {which indicates the existence of intersectional bias}. Similarly in class Real also, pairs involving the light skin female group show bias with very low values for DP and even EO. That is, pairwise bias analysis measures could unveil \hl{the} existence of bias in the \{Light skin Female vs. Light skin Male\}, \{Dark skin Female vs. Light skin Female\} and \{Light skin Female vs. Dark skin Male\} intersectional pairs. 
The prediction intensity (confidence) plots of the images of intersectional pairs in the bias evaluation corpora that \hl{show} unbiased and biased results are given in figure \ref{fig_3bias_result_vit_unbiased} and \ref{fig_3bias_result_vit_biased}, respectively. The horizontal axes of the plots indicate images in the bias evaluation corpora (that contains a total of 500 GAN/Real images) and the vertical axis indicates the prediction intensity score of each of these images. For the unbiassed intersectional pair \{Dark skin Male vs. Light skin Male\} (in figure \ref{fig_3bias_result_vit_unbiased}), it can be observed that there is not much difference in prediction intensities within the pairs for both GAN and Real classes. Whereas, in figure \ref{fig_3bias_result_vit_biased}, the comparatively much \hl{greater} difference in prediction intensities within the pairs for both GAN and Real classes, subsidize the quantitative results in table \ref{table_3bias_vit} that indicates the existence of bias in the intersectional pair \{Light skin Female vs. Dark skin Male\}.

\subsubsection{CvT}

The bias evaluation results of the CvT based model for various domains \hl{are} shown in table \ref{table_3bias_cvt}. From the top portion of the table showing the results of individual measures, it can be observed that CvT shows high and similar accuracies for all the social groups within each of the domains. The FPR and FNR values are also very low and similar across the social groups within each domain. The bottom portion of the same table \ref{table_3bias_cvt} presents the results of pairwise bias analysis of the model for various domains. The measures DP and EO report very high values, nearly similar to an ideal unbiased scenario. Altogether, the individual and pairwise measures do not report any existence of significant bias in the CvT based model.

\begin{table}[!h]
\centering
\caption{Evaluation results of \textbf{CvT} in \textbf{uncompressed} setting}
\label{table_3bias_cvt}
\resizebox{\linewidth}{!}{
\setlength{\tabcolsep}{3pt}
\begin{tabular}{@{}c|cc|cc|cc|cccc@{}}
\toprule
\multicolumn{10}{c}{\textbf{Individual measures based analysis}} \\
\midrule
\multicolumn{1}{c|}{\multirow{2}{*}{\hl{Measure}}} &
  \multicolumn{2}{c|}{Gender} &
  \multicolumn{2}{c|}{Race} &
  \multicolumn{2}{c|}{Affective} &
  \multicolumn{4}{c}{Intersection} \\ \cmidrule(l){2-11} 
\multicolumn{1}{c|}{} &
  \multicolumn{1}{c|}{F} &
  \multicolumn{1}{c|}{M} &
  \multicolumn{1}{c|}{D} &
  \multicolumn{1}{c|}{L} &
  \multicolumn{1}{c|}{Ns} &
  \multicolumn{1}{c|}{S} &
  \multicolumn{1}{c|}{D+F} &
  \multicolumn{1}{c|}{D+M} &
  \multicolumn{1}{c|}{L+F} &
  \multicolumn{1}{c}{L+M} \\ \midrule
Acc & 99.05 & 99.15 & 98.80  & 99.40  & 99.45 & 99.40  & 98.80  & 98.80  & 99.30  & 99.50  \\ \midrule
Acc\textsubscript{gan} & 98.40   & 98.60  & 98.10  & 98.90  & 99.10  & 99.20 & 98.00  & 98.20  & 98.80  & 99.00  \\ \midrule
Acc\textsubscript{real} & 99.70   & 99.70  & 99.50  & 99.90  & 99.80  & 99.60  & 99.60  & 99.40  & 99.80  & 100.0  \\ \midrule
FPR & 0.003  & 0.003 & 0.005 & 0.001 & 0.002 & 0.004 & 0.004 & 0.006 & 0.002 & 0.000 \\ \midrule
FNR & 0.016  & 0.014 & 0.019 & 0.011 & 0.009 & 0.008 & 0.020  & 0.018 & 0.012 & 0.010 \vspace{5pt}\\ %\bottomrule
\end{tabular}}
\resizebox{\linewidth}{!}{
\setlength{\tabcolsep}{1.5pt}
\begin{tabular}{@{}c|c|c|c|cccccc@{}}
\toprule
\multicolumn{10}{c}{\textbf{Pairwise measures based analysis}} \\
\midrule
\multicolumn{1}{c|}{} & \multicolumn{1}{c|}{Gender} & \multicolumn{1}{c|}{Race} & \multicolumn{1}{c|}{Affect} & \multicolumn{6}{c}{Intersection} \\ \midrule
\multicolumn{1}{c|}{} &
  \multicolumn{1}{c|}{F$\times$M} &
  \multicolumn{1}{c|}{D$\times$L} &
  \multicolumn{1}{c|}{Ns$\times$S} &
  \multicolumn{1}{c|}{\begin{tabular}[c]{@{}l@{}} D+F $\times$ \\ D+M \end{tabular}} &
  \multicolumn{1}{c|}{\begin{tabular}[c]{@{}l@{}} L+F $\times$ \\ L+M \end{tabular}} &
  \multicolumn{1}{c|}{\begin{tabular}[c]{@{}l@{}} D+F $\times$ \\ L+F \end{tabular}} &
  \multicolumn{1}{c|}{\begin{tabular}[c]{@{}l@{}} D+M $\times$ \\ L+M \end{tabular}} &
  \multicolumn{1}{c|}{\begin{tabular}[c]{@{}l@{}} D+F $\times$ \\ L+M \end{tabular}} &
  \multicolumn{1}{c}{\begin{tabular}[c]{@{}l@{}} L+F $\times$ \\ D+M \end{tabular}}  \\ \midrule
\multicolumn{10}{c}{GAN} \\
\midrule
ACS & +0.0050 & +0.0151 & -0.0010 & +0.0064 & +0.0036 & +0.0165 & +0.0138 & +0.0200 & -0.0103 \\ \midrule
DP  & 0.9980  & 0.9960  & 0.9970  & 0.9960  & 1.0000   & 0.9939  & 0.9980  & 0.9939  & 0.9980  \\ \midrule
EO  & 0.9980  & 0.9919  & 0.9990  & 0.9980  & 0.9980  & 0.9919  & 0.9919  & 0.9899  & 0.9939  \\ \midrule
\multicolumn{10}{c}{Real} \\ \midrule
ACS & -0.0005 & +0.0063 & +0.0012 & -0.0004 & -0.0007 & +0.0064 & +0.0062 & +0.0058 & -0.0068 \\ \midrule
DP  & 0.9980  & 0.9961  & 0.9970  & 0.9961  & 1.0000   & 0.9941  & 0.9980  & 0.9941  & 0.9980  \\ \midrule
EO  & 1.0000  & 0.9960  & 0.9980  & 0.9980  & 0.9980  & 0.9980  & 0.9940  & 0.9960  & 0.9960  \\ \bottomrule
\end{tabular}}
\end{table}

\subsubsection{Swin transformer}

The bias evaluation results of the Swin transformer based model for various domains \hl{are} shown in table \ref{table_3bias_swin}. Similar to the CvT based model, it can be observed from the top portion of the table with individual measures that, \hl{the} Swin transformer shows high and similar accuracies for all categories of social groups. The FPR and FNR values are also very low and similar across the social groups within each domain. \hl{The} bottom portion of the same table \ref{table_3bias_swin} presents the results of pairwise bias analysis of the model for various domains. The measures DP and EO \hl{show} very high values, nearly similar to an ideal unbiased scenario. Altogether, the individual and pairwise measures do not report any existence of significant bias in the Swin transformer model.

\begin{table}[!h]
\centering
\caption{Evaluation results of \textbf{Swin} transformer in \textbf{uncompressed} setting}
\label{table_3bias_swin}
\resizebox{\linewidth}{!}{
\setlength{\tabcolsep}{3pt}
\begin{tabular}{@{}c|cc|cc|cc|cccc@{}}
\toprule
\multicolumn{10}{c}{\textbf{Individual measures based analysis}} \\
\midrule
\multicolumn{1}{c|}{\multirow{2}{*}{\hl{Measure}}} &
  \multicolumn{2}{c|}{Gender} &
  \multicolumn{2}{c|}{Race} &
  \multicolumn{2}{c|}{Affective} &
  \multicolumn{4}{c}{Intersection} \\ \cmidrule(l){2-11} 
\multicolumn{1}{c|}{} &
  \multicolumn{1}{c|}{F} &
  \multicolumn{1}{c|}{M} &
  \multicolumn{1}{c|}{D} &
  \multicolumn{1}{c|}{L} &
  \multicolumn{1}{c|}{Ns} &
  \multicolumn{1}{c|}{S} &
  \multicolumn{1}{c|}{D+F} &
  \multicolumn{1}{c|}{D+M} &
  \multicolumn{1}{c|}{L+F} &
  \multicolumn{1}{c}{L+M} \\ \midrule
Acc & 99.30  & 99.95 & 99.70  & 99.55 & 99.80  & 99.40  & 99.40  & 100.0 & 99.20  & 99.90  \\ \midrule
Acc\textsubscript{gan} & 99.40  & 100.0   & 99.60  & 99.80  & 99.80  & 99.60  & 99.20  & 100.0   & 99.60  & 100.0  \\ \midrule
Acc\textsubscript{real} & 99.20  & 99.90  & 99.80  & 99.30  & 99.80  & 99.20  & 99.60  & 100.0   & 98.80  & 99.80  \\ \midrule
FPR  & 0.008 & 0.001 & 0.002 & 0.007 & 0.002 & 0.008 & 0.004 & 0.000 & 0.012 & 0.002 \\ \midrule
FNR  & 0.006 & 0.000 & 0.004 & 0.002 & 0.002 & 0.004 & 0.008 & 0.000 & 0.004 & 0.000 \vspace{5pt}\\ %\bottomrule
\end{tabular}}
\resizebox{\linewidth}{!}{
\setlength{\tabcolsep}{1.5pt}
\begin{tabular}{@{}c|c|c|c|cccccc@{}}
\toprule
\multicolumn{10}{c}{\textbf{Pairwise measures based analysis}} \\
\midrule
\multicolumn{1}{c|}{} & \multicolumn{1}{c|}{Gender} & \multicolumn{1}{c|}{Race} & \multicolumn{1}{c|}{Affect} & \multicolumn{6}{c}{Intersection} \\ \midrule
\multicolumn{1}{c|}{} &
  \multicolumn{1}{c|}{F$\times$M} &
  \multicolumn{1}{c|}{D$\times$L} &
  \multicolumn{1}{c|}{Ns$\times$S} &
  \multicolumn{1}{c|}{\begin{tabular}[c]{@{}l@{}} D+F $\times$ \\ D+M \end{tabular}} &
  \multicolumn{1}{c|}{\begin{tabular}[c]{@{}l@{}} L+F $\times$ \\ L+M \end{tabular}} &
  \multicolumn{1}{c|}{\begin{tabular}[c]{@{}l@{}} D+F $\times$ \\ L+F \end{tabular}} &
  \multicolumn{1}{c|}{\begin{tabular}[c]{@{}l@{}} D+M $\times$ \\ L+M \end{tabular}} &
  \multicolumn{1}{c|}{\begin{tabular}[c]{@{}l@{}} D+F $\times$ \\ L+M \end{tabular}} &
  \multicolumn{1}{c}{\begin{tabular}[c]{@{}l@{}} L+F $\times$ \\ D+M \end{tabular}}  \\ \midrule
\multicolumn{10}{c}{GAN} \\
\midrule
ACS &  +0.0026 & +0.0034 & -0.0002 & +0.0059 & -0.0007 & +0.0067 & +0.0002 & +0.0060 & -0.0009 \\ \midrule
DP  & 0.9990  & 0.9930  & 0.9960  & 0.9960   & 0.9940  & 0.9881  & 0.9980  & 0.9940  & 0.9921  \\ \midrule
EO  & 0.9940   & 0.9980  & 0.9980  & 0.9920  & 0.9960  & 0.9960  & 1.0000  & 0.9920  & 0.9960  \\ \midrule
\multicolumn{10}{c}{Real} \\ \midrule
ACS & -0.0014 & +0.0010 & +0.0008 & -0.0006 & -0.0022 & +0.0018 & +0.0002 & -0.0004 & -0.0024 \\ \midrule
DP  & 0.9990  & 0.9930  & 0.9960   & 0.9960  & 0.9940  & 0.9880  & 0.9980   & 0.9940  & 0.9920  \\ \midrule
EO  & 0.9930  & 0.9950  & 0.9940  & 0.9960   & 0.9900  & 0.9920  & 0.9980   & 0.9980  & 0.9980  \\ \bottomrule
\end{tabular}}
\end{table}

\subsection{Bais analysis in the \textbf{compressed} evaluation setting}
\label{sec_3bias_results_compressed}

\subsubsection{ViT}

The results of individual and pairwise measures of bias analysis of the ViT based model on the JPEG compressed evaluation corpora \hl{are} shown in table \ref{table_3bias_vit_compressed}. Similar to the observations discussed in the study \cite{gangan2023robust}, it can be observed that accuracies of the model \hl{decrease} over compressed images, and the decrease in accuracy is much higher for the class GAN than class Real. In the compressed evaluation setting of ViT, it can also be observed that the difference in GAN accuracies (Acc\textsubscript{gan}) between the groups within each of the domains has increased than its corresponding uncompressed evaluation setting. For example, in the uncompressed evaluation setting of ViT, the difference in GAN accuracies between female and male groups within the gender domain is 0.4 percentage points (in table \ref{table_3bias_vit}); But, in this compressed setting, the difference has increased to 1.5 percentage points. Similarly, 2.8 percentage points of difference in racial domain between dark skin and light skin groups in the uncompressed evaluation setting have increased to 5.1 percentage points in this compressed setting, and 1.2 percentage points of difference between non-smiling and smiling groups of affective domain have increased to 3.2 percentage points. Similar to the uncompressed setting, here also in the class GAN, for the intersectional domain, light skin female group obtains the highest accuracy and dark skin female obtains the lowest accuracy, but \hl{the} difference in their accuracies (Acc\textsubscript{gan}) increases to 11.4 percentage points compared to 6.6 percentage points in the previous uncompressed setting. Altogether, for the class GAN, gender bias against the male group (lower accuracy for male than female), racial bias against dark skin, affective bias against non-smiling group, and intersectional biases, particularly against dark skin female, has increased in the compressed evaluation setting when compared to the uncompressed evaluation setting. Also, similar to the uncompressed setting, this compressed setting \hl{has} higher values of FPR for female group in gender domain, light skin in racial domain, smiling face group in affective doman, and light skin females in the intersectional domain indicating that Real images of these groups are highly likely to be misclassified as GAN images.

\begin{table}[!h]
\centering
\caption{Evaluation results of \textbf{ViT} in \textbf{compressed} setting}
\label{table_3bias_vit_compressed}
\resizebox{\linewidth}{!}{
\setlength{\tabcolsep}{3pt}
\begin{tabular}{@{}c|cc|cc|cc|cccc@{}}
\toprule
\multicolumn{10}{c}{\textbf{Individual measures based analysis}} \\
\midrule
\multicolumn{1}{c|}{\multirow{2}{*}{\hl{Measure}}} &
  \multicolumn{2}{c|}{Gender} &
  \multicolumn{2}{c|}{Race} &
  \multicolumn{2}{c|}{Affective} &
  \multicolumn{4}{c}{Intersection} \\ \cmidrule(l){2-11} 
\multicolumn{1}{c|}{} &
  \multicolumn{1}{c|}{F} &
  \multicolumn{1}{c|}{M} &
  \multicolumn{1}{c|}{D} &
  \multicolumn{1}{c|}{L} &
  \multicolumn{1}{c|}{Ns} &
  \multicolumn{1}{c|}{S} &
  \multicolumn{1}{c|}{D+F} &
  \multicolumn{1}{c|}{D+M} &
  \multicolumn{1}{c|}{L+F} &
  \multicolumn{1}{c}{L+M} \\ \midrule
Acc & 85.90  & 89.10  & 89.15 & 85.85 & 89.40  & 88.40  & 87.30  & 91.00  & 84.50  & 87.20  \\ \midrule
Acc\textsubscript{gan} & 86.10  & 84.60  & 82.80  & 87.90  & 86.00  & 89.20  & 80.40  & 85.20  & 91.80  & 84.00  \\ \midrule
Acc\textsubscript{real} & 85.70  & 93.60  & 95.50  & 83.80  & 92.80  & 87.60  & 94.20  & 96.80  & 77.20  & 90.40  \\ \midrule
FPR & 0.143 & 0.064 & 0.045 & 0.162 & 0.072 & 0.124 & 0.058 & 0.032 & 0.228 & 0.096 \\ \midrule
FNR & 0.139 & 0.154 & 0.172 & 0.121 & 0.140 & 0.108 & 0.196 & 0.148 & 0.082 & 0.160 \vspace{5pt}\\ %\bottomrule
\end{tabular}}
\resizebox{\linewidth}{!}{
\setlength{\tabcolsep}{1.5pt}
\begin{tabular}{@{}c|c|c|c|cccccc@{}}
\toprule
\multicolumn{10}{c}{\textbf{Pairwise measures based analysis}} \\
\midrule
\multicolumn{1}{c|}{} & \multicolumn{1}{c|}{Gender} & \multicolumn{1}{c|}{Race} & \multicolumn{1}{c|}{Affect} & \multicolumn{6}{c}{Intersection} \\ \midrule
\multicolumn{1}{c|}{} &
  \multicolumn{1}{c|}{F$\times$M} &
  \multicolumn{1}{c|}{D$\times$L} &
  \multicolumn{1}{c|}{Ns$\times$S} &
  \multicolumn{1}{c|}{\begin{tabular}[c]{@{}l@{}} D+F $\times$ \\ D+M \end{tabular}} &
  \multicolumn{1}{c|}{\begin{tabular}[c]{@{}l@{}} L+F $\times$ \\ L+M \end{tabular}} &
  \multicolumn{1}{c|}{\begin{tabular}[c]{@{}l@{}} D+F $\times$ \\ L+F \end{tabular}} &
  \multicolumn{1}{c|}{\begin{tabular}[c]{@{}l@{}} D+M $\times$ \\ L+M \end{tabular}} &
  \multicolumn{1}{c|}{\begin{tabular}[c]{@{}l@{}} D+F $\times$ \\ L+M \end{tabular}} &
  \multicolumn{1}{c}{\begin{tabular}[c]{@{}l@{}} L+F $\times$ \\ D+M \end{tabular}}  \\ \midrule
\multicolumn{10}{c}{GAN} \\ \midrule
ACS & -0.0068 & +0.0157 & +0.0066 & +0.0212 & -0.0323 & +0.0411 & -0.0113 & +0.0101 & -0.0208 \\ \midrule
DP  & 0.9064  & 0.8386  & 0.9173  & 0.9751  & 0.8168  & \textbf{0.7522}  & 0.9444  & 0.9209  & \textbf{0.7714}  \\ \midrule
EO  & 0.9826  & 0.9420  & 0.9641  & 0.9437  & 0.9150  & 0.8758  & 0.9859  & 0.9571  & 0.9281  \\ \midrule
\multicolumn{10}{c}{Real} \\ \midrule
ACS & +0.0165 & -0.0121 & -0.0168 & +0.0129 & +0.0216 & -0.0174 & -0.0085 & +0.0045 & 0.0298 \\ \midrule
DP  & 0.9138  & 0.8509  & 0.9214  & 0.9807  & \textbf{0.8026}  & \textbf{0.7504}  & 0.9534  & 0.9350  & \textbf{0.7652}  \\ \midrule
EO  & 0.9156  & 0.8775  & 0.944   & 0.9731  & 0.8540  & 0.8195  & 0.9339  & 0.9597  & 0.7975  \\ \bottomrule
\end{tabular}}
\end{table}

\hl{The} bottom portion of the table \ref{table_3bias_vit_compressed} presents the results of pairwise measures of bias analysis of ViT based model over the compressed evaluation corpora. The results show that, in this compressed setting, for class GAN, there is a decrease in DP and EO values when compared to the corresponding uncompressed evaluation setting. For example, the DP of \{Dark skin vs. Light skin\} for class GAN has decreased from 0.8758 (in previous uncompressed evaluation setting, table \ref{table_3bias_vit}) to 0.8386 (in current compressed evaluation setting, table \ref{table_3bias_vit_compressed}), DP of \{Dark skin Female vs Light skin Female\} has decreased from 0.7989 to 0.7522, etc. Thus, the pairwise evaluations on ViT also show that, for class GAN, the biases increase over the compressed images than the uncompressed images. That is, this indicates biases in the class GAN gets amplified with compression.

\subsubsection{CvT}

The results of individual and pairwise measures of bias analysis of the CvT based model over the compressed evaluation corpora \hl{are} shown in table \ref{table_3bias_cvt_compressed}. The top portion of the table shows \hl{the} results of individual measures. Similar to the results of ViT (in table \ref{table_3bias_vit_compressed}), compression decreases accuracies of the CvT based model, particularly the class GAN accuracy (Acc\textsubscript{gan}), whereas class Real (Acc\textsubscript{real}) maintains its high accuracies. But compared to ViT, the drop in the accuracies for class GAN of the CvT based model is massively very high. Also, this accuracy decay in CvT is not similar across different social groups within a domain, indicating high bias. 

\begin{table}[!h]
\centering
\caption{Results of individual measures evaluating \textbf{CvT} in \textbf{compressed} setting}
\label{table_3bias_cvt_compressed}
\resizebox{\linewidth}{!}{
\setlength{\tabcolsep}{3pt}
\begin{tabular}{@{}c|cc|cc|cc|cccc@{}}
\toprule
\multicolumn{10}{c}{\textbf{Individual measures based analysis}} \\
\midrule
\multicolumn{1}{c|}{\multirow{2}{*}{\hl{Measure}}} &
  \multicolumn{2}{c|}{Gender} &
  \multicolumn{2}{c|}{Race} &
  \multicolumn{2}{c|}{Affective} &
  \multicolumn{4}{c}{Intersection} \\ \cmidrule(l){2-11} 
\multicolumn{1}{c|}{} &
  \multicolumn{1}{c|}{F} &
  \multicolumn{1}{c|}{M} &
  \multicolumn{1}{c|}{D} &
  \multicolumn{1}{c|}{L} &
  \multicolumn{1}{c|}{Ns} &
  \multicolumn{1}{c|}{S} &
  \multicolumn{1}{c|}{D+F} &
  \multicolumn{1}{c|}{D+M} &
  \multicolumn{1}{c|}{L+F} &
  \multicolumn{1}{c}{L+M} \\ \midrule
Acc & 51.65 & 51.15 & 51.05 & 51.75 & 51.70  & 52.05 & 50.80  & 51.30 & 52.50 & 51.00  \\ \midrule
Acc\textsubscript{gan} & 34.00  & 23.00  & 22.00  & 35.00  & 34.00  & 42.00  & 1.80  & 2.60   & 5.00  & 2.00   \\ \midrule
Acc\textsubscript{real} & 99.90  & 100.0   & 99.90  & 100.0   & 100.0   & 99.90  & 99.80  & 100.0   & 100.0  & 100.0  \\ \midrule
FPR & 0.001 & 0.000 & 0.001 & 0.000 & 0.000 & 0.001 & 0.002 & 0.000 & 0.000 & 0.000 \\ \midrule
FNR & 0.966 & 0.977 & 0.978 & 0.965 & 0.966 & 0.958 & 0.982 & 0.974 & 0.950 & 0.980 \vspace{5pt}\\ %\bottomrule
\end{tabular}}
\resizebox{\linewidth}{!}{
\setlength{\tabcolsep}{1.5pt}
\begin{tabular}{@{}c|c|c|c|cccccc@{}}
\toprule
\multicolumn{10}{c}{\textbf{Pairwise measures based analysis}} \\
\midrule
\multicolumn{1}{c|}{} & \multicolumn{1}{c|}{Gender} & \multicolumn{1}{c|}{Race} & \multicolumn{1}{c|}{Affect} & \multicolumn{6}{c}{Intersection} \\ \midrule
\multicolumn{1}{c|}{} &
  \multicolumn{1}{c|}{F$\times$M} &
  \multicolumn{1}{c|}{D$\times$L} &
  \multicolumn{1}{c|}{Ns$\times$S} &
  \multicolumn{1}{c|}{\begin{tabular}[c]{@{}l@{}} D+F $\times$ \\ D+M \end{tabular}} &
  \multicolumn{1}{c|}{\begin{tabular}[c]{@{}l@{}} L+F $\times$ \\ L+M \end{tabular}} &
  \multicolumn{1}{c|}{\begin{tabular}[c]{@{}l@{}} D+F $\times$ \\ L+F \end{tabular}} &
  \multicolumn{1}{c|}{\begin{tabular}[c]{@{}l@{}} D+M $\times$ \\ L+M \end{tabular}} &
  \multicolumn{1}{c|}{\begin{tabular}[c]{@{}l@{}} D+F $\times$ \\ L+M \end{tabular}} &
  \multicolumn{1}{c}{\begin{tabular}[c]{@{}l@{}} L+F $\times$ \\ D+M \end{tabular}}  \\ \midrule
\multicolumn{10}{c}{GAN} \\
\midrule
ACS & -0.0267 & +0.0591 & -0.0331 & -0.0886 & +0.0432 & -0.0012 & +0.1200 & +0.0420 & -0.0872 \\ \midrule
DP  & \textbf{0.6571}  & \textbf{0.6571}  & \textbf{0.7907}  & \textbf{ 0.7692}  & \textbf{0.3999}  & \textbf{0.3999}  & \textbf{0.7692}  & 1.0000   & \textbf{0.5199} \\ \midrule
EO  & 0.6765  & 0.6280  & 0.8095  & 0.6923  & 0.3999  & 0.3599  & 0.7692  & 0.8999  & 0.5199  \\ \midrule
\multicolumn{10}{c}{Real} \\ \midrule
ACS & +0.0027 & -0.0021 & -0.0027 & +0.0024 & +0.0029 & -0.0024 & -0.0019 & +0.0005 & 0.0048 \\ \midrule
DP  & 0.9939  & 0.9939  & 0.9954  & 0.9970  & 0.9849  & 0.9849  & 0.9970  & 1.0000     & 0.9878  \\ \midrule
EO  & 0.9990   & 0.9990   & 0.9990   & 0.9980   & 1.0000   & 0.9980   & 1.0000    & 0.9980   & 1.0000  \\ \bottomrule
\end{tabular}}
\end{table}

\hl{The} bottom portion of the table \ref{table_3bias_cvt_compressed} presents the results of pairwise analysis of CvT over the compressed evaluation corpora. From the table, it can be understood that, for the class GAN predictions of CvT, the ideal unbiased scenario which was seen in the previous uncompressed evaluation setting of CvT (in table \ref{table_3bias_cvt}) has been completely overturned to a very largely biased scenario due to compression. This is because the drop in GAN accuracies are not similar across the groups within a domain (except for the dark skin female vs. light skin male pairs). Whereas, it can be observed that the class Real predictions of the CvT still maintains the ideal unbiased scenario as in the previous uncompressed evaluation setting.

\subsubsection{Swin transformer}

The results of individual and pairwise measures of bias analysis of the Swin transformer based model over the compressed evaluation corpora \hl{are} shown in table \ref{table_3bias_swin_compressed}. The top portion of the table shows the results of individual measures. In this model also the GAN accuracy (Acc\textsubscript{gan}) decreases due to compression, thereby decreasing the total model accuracy. Contrary to the previous uncompressed setting of Swin transformer (in table \ref{table_3bias_swin}), where similar and high accuracies are obtained for all social groups within each domain, this compressed evaluation setting has eventually brought up differences in GAN accuracies across social groups within each of the domains. That is, the GAN accuracy of the male group is less than the female group by 2.5 percentage points in the gender domain, the dark skin group is less than the light skin group by 4.9 percentage points in the racial domain, and the non-smiling group is less than smiling group by 3.2 percentage points in the affective domain. In the intersectional domain, the highest GAN accuracy is obtained for the light skin female group and lowest for the dark skin female group, a very high accuracy difference of 16.4 percentage points is observed between these two intersectional groups for class GAN. Thus these accuracy differences indicate {the existence of} bias in the compressed setting for the class GAN of Swin transformer based model.

\begin{table}[!h]
\centering
\caption{Evaluation results of \textbf{Swin} transformer in \textbf{compressed} setting}
\label{table_3bias_swin_compressed}
\resizebox{\linewidth}{!}{
\setlength{\tabcolsep}{3pt}
\begin{tabular}{@{}c|cc|cc|cc|cccc@{}}
\toprule
\multicolumn{10}{c}{\textbf{Individual measures based analysis}} \\
\midrule
\multicolumn{1}{c|}{\multirow{2}{*}{\hl{Measure}}} &
  \multicolumn{2}{c|}{Gender} &
  \multicolumn{2}{c|}{Race} &
  \multicolumn{2}{c|}{Affective} &
  \multicolumn{4}{c}{Intersection} \\ \cmidrule(l){2-11} 
\multicolumn{1}{c|}{} &
  \multicolumn{1}{c|}{F} &
  \multicolumn{1}{c|}{M} &
  \multicolumn{1}{c|}{D} &
  \multicolumn{1}{c|}{L} &
  \multicolumn{1}{c|}{Ns} &
  \multicolumn{1}{c|}{S} &
  \multicolumn{1}{c|}{D+F} &
  \multicolumn{1}{c|}{D+M} &
  \multicolumn{1}{c|}{L+F} &
  \multicolumn{1}{c}{L+M} \\ \midrule
Acc & 83.10  & 82.45 & 81.95 & 83.60  & 84.39 & 85.70  & 79.60 & 84.30  & 86.60  & 80.60   \\ \midrule
Acc\textsubscript{gan} & 68.20  & 65.70  & 64.50  & 69.40  & 69.60  & 72.80  & 60.00  & 69.00  & 76.40  & 62.40   \\ \midrule
Acc\textsubscript{real} & 98.00  & 99.20  & 99.40  & 97.80  & 99.20  & 98.60  & 99.20  & 99.60  & 96.80  & 98.80  \\ \midrule
FPR & 0.020  & 0.008 & 0.006 & 0.022 & 0.008 & 0.014 & 0.008 & 0.004 & 0.032 & 0.012 \\ \midrule
FNR & 0.318 & 0.343 & 0.355 & 0.306 & 0.304 & 0.272 & 0.40   & 0.310  & 0.236 & 0.376 \vspace{5pt}\\ %\bottomrule
\end{tabular}}
\resizebox{\linewidth}{!}{
\setlength{\tabcolsep}{1.5pt}
\begin{tabular}{@{}c|c|c|c|cccccc@{}}
\toprule
\multicolumn{10}{c}{\textbf{Pairwise measures based analysis}} \\
\midrule
\multicolumn{1}{c|}{} & \multicolumn{1}{c|}{Gender} & \multicolumn{1}{c|}{Race} & \multicolumn{1}{c|}{Affect} & \multicolumn{6}{c}{Intersection} \\ \midrule
\multicolumn{1}{c|}{} &
  \multicolumn{1}{c|}{F$\times$M} &
  \multicolumn{1}{c|}{D$\times$L} &
  \multicolumn{1}{c|}{Ns$\times$S} &
  \multicolumn{1}{c|}{\begin{tabular}[c]{@{}l@{}} D+F $\times$ \\ D+M \end{tabular}} &
  \multicolumn{1}{c|}{\begin{tabular}[c]{@{}l@{}} L+F $\times$ \\ L+M \end{tabular}} &
  \multicolumn{1}{c|}{\begin{tabular}[c]{@{}l@{}} D+F $\times$ \\ L+F \end{tabular}} &
  \multicolumn{1}{c|}{\begin{tabular}[c]{@{}l@{}} D+M $\times$ \\ L+M \end{tabular}} &
  \multicolumn{1}{c|}{\begin{tabular}[c]{@{}l@{}} D+F $\times$ \\ L+M \end{tabular}} &
  \multicolumn{1}{c}{\begin{tabular}[c]{@{}l@{}} L+F $\times$ \\ D+M \end{tabular}}  \\ \midrule
\multicolumn{10}{c}{GAN} \\
\midrule
ACS & -0.0008 & -0.0028 & -0.0020 & +0.0242 & -0.0254 & +0.0213 & -0.0284 & -0.0035 & 0.0029 \\ \midrule
DP  & 0.9473  & 0.9092  & 0.9488  & 0.8761  & \textbf{0.7989}  & \textbf{0.7638}  & 0.9164  & 0.9559  & 0.8718 \\ \midrule
EO  & 0.9633  & 0.9294  & 0.9560  & 0.8695  & 0.8167  & 0.7853  & 0.9043  & 0.9615  & 0.9031  \\ \midrule
\multicolumn{10}{c}{Real} \\ \midrule
ACS & +0.0025 & -0.0057 & -0.0016 & -0.0006 & +0.0057 & -0.0089 & -0.0026 & -0.0032 & 0.0083 \\ \midrule
DP & 0.9723  & 0.9518  & 0.9707  & 0.9382  & 0.8826  & 0.8649  & 0.9574  & 0.9798  & 0.9218  \\ \midrule
EO  & 0.9879  & 0.9839  & 0.9940  & 0.9959  & 0.9797  & 0.9758  & 0.9919  & 0.9959  & 0.9718  \\ \bottomrule
\end{tabular}}
\end{table}

\hl{The} bottom portion of the table \ref{table_3bias_swin_compressed} presents the results of pairwise analysis of the Swin transformer based model over the compressed evaluation corpora. Compared to the previous uncompressed setting of Swin transformer (in table \ref{table_3bias_swin}) that reports nearly an ideal unbiased scenario, in this compressed setting, values of the measures DP and EO \hl{decrease} highly for the class GAN indicating an increase in bias for the class GAN predictions. A high bias for class GAN predictions can be observed particularly in the pairs, light skin female vs. light skin male and dark skin female vs. light skin female.

\section{{Discussion}}
The bias evaluation results \hl{show} that in uncompressed evaluation settings, the evaluation corpora and measures could identify bias in the ViT based forensic classifier model, such as, bias in pairs involving light skin female groups e.g., bias in light skin female vs. dark skin male, dark skin female vs. light skin female, etc. Also, bias analysis in ViT based model shows other interesting inferences such as, Real images of light skin females have a very high probability of being misclassified as GAN images, and GAN images of dark skin females have a very high probability of being misclassified as Real images. However, the uncompressed setting could not identify any bias in the CvT and the Swin transformer based models. 

The compressed evaluation setting, on the other hand, identifies high bias in all three transformer based models, particularly in the class GAN predictions. In the compressed evaluation setting, for the class GAN predictions of all the models, gender bias against the male group, racial bias against dark skin group, affective bias against non-smiling group, and intersectional biases, particularly against dark skin female group, has increased when compared to the uncompressed evaluation setting. Bias is identified in all the domains for the CvT based model, in the compressed evaluation setting. That is, the study could observe that model bias is impacted by image compression. Moreover, the model bias identified in the uncompressed setting is observed to be amplified in the compressed setting, particularly for the class GAN predictions. Also, given that our results indicate Real images of certain social groups such as light skin females are more likely to be misclassified as GAN images, and GAN images of dark skin females are more likely to be misclassified as Real images, etc., the images of these social groups when compressed can even more increase the risk of security threats. Therefore, image forensics works that utilize visual transformers for the task of distinguishing natural and GAN generated images, besides assessing the robustness of the algorithms towards image compression, should also study the existence of bias in these algorithms, even in the compressed setting.

ViT and Swin transformer based models chosen for this study are pre-trained on the ImageNet-21K dataset \cite{ridnik2021imagenet}. As already stated above, more than the uncompressed evaluation settings, these models show a higher bias in their corresponding compressed evaluation settings. On the other hand, the model CvT is pre-trained on the ImageNet-1k dataset \cite{russakovsky2015imagenet}. But unlike ViT and Swin transformer, CvT has comparatively a very high transition from an ideal unbiased scenario in the uncompressed evaluation setting to a very largely biased model in the compressed evaluation setting. Hence, pre-training corpora of the visual transformers might be one of the factors inducing bias in these models.

\section{Conclusion}
\label{sec_3bias_conclu}

This study explored bias in {the visual transformer based} image forensic algorithms that classify natural and GAN generated images. The study utilized three visual transformers viz., ViT, CvT, and Swin, for constructing image forensic algorithms to classify natural and GAN images. The pre-trained visual transformers are fine-tuned using the task-specific natural image versus GAN image dataset, and are examined for any existence of bias in the gender, racial, affective, and even intersectional domains. Hence, a bias evaluation corpora consisting of social groups belonging to the evaluation domains are procured for the study. Individual and pairwise bias evaluation measures are used for identifying any existence of bias in these transformer based forensic models. Since robustness towards image compression is significant for the forensic algorithms, this study also examines the role of image compression on model bias. To the best knowledge, this is the first work to study the impact of image compression on model bias, particularly focusing on the task of classifying natural and GAN images. Hence, this work conducts the bias evaluation experiments in two separate settings; one set of experiments on the original uncompressed evaluation corpora and the other on the compressed version of the same evaluation corpora, where both these experiments rely on same evaluation measures. 

This study helped to identify the existence of bias in the transformer based models for the task of distinguishing natural and GAN generated images. The two-phase bias evaluation strategy helped to \hl{identify} bias in the uncompressed and compressed scenarios and also to study the impact of image compression on the model bias. The study observed that image compression impacts model biases, and particularly compression amplifies the biases of the class GAN predictions. To help towards future research, all relevant materials of this study including the source codes will be made publicly available at \url{https://github.com/manjaryp/ImageForgeryFairness} and \url{https://dcs.uoc.ac.in/cida/projects/dif/Imageforgeryfairness.html} along with the publication.

\subsection{\hl{Future directions}}
\label{sec_3bias_future_directions}

\hl{This study to explore algorithmic fairness in image forensic systems proposes a generalized and simple bias evaluation framework consisting of evaluation domains, evaluation corpora, evaluation measures, and also the two evaluation settings, one in uncompressed and the other in compressed settings. Hence, beyond the popular transformers such as ViT, CvT, and Swin analyzed for algorithmic fairness in this work, our generalized and simple bias evaluation framework makes it highly suitable for easily evaluating the latest visual transformers such as RT-DETR \mbox{\cite{lv2023detrs}}, Conv2Former \mbox{\cite{hou2024conv}}, etc. Also, the results of our bias analysis experiments could observe that the differences in the existence of biases in corresponding uncompressed and compressed settings have similar trends for visual transformers pre-trained on the same dataset. For example, the transition in bias existences from the uncompressed to compressed settings of ViT and Swin transformers pre-trained on ImageNet-21K are similar, whereas the transition is different for CvT pre-trained on the ImageNet-1k dataset. Therefore, this analysis from our study could be useful in the future for exploring algorithmic bias in many other latest visual transformers such as Conv2Former-B \mbox{\cite{hou2024conv}} utilizing the pre-trained dataset ImageNet-21K or MobileViT \mbox{\cite{mehta2021mobilevit}} utilizing ImageNet-1k dataset, to further expand the study particularly on the cause and relationship with pre-training data biases.} 

In the future, \hl{we are considering} to extend this work to analyze various factors that cause or originate these biases. Although it is cumbersome to procure datasets with balanced groups within a wide variety of domains, this could, in the future, help in determining various sources of biases, especially in identifying the existence of any pre-train and \hl{fine-tuning} data biases. \hl{The evaluation corpora can also be expanded and annotated towards a benchmark corpora to explore bias in many other domains, in the context of image forensics.} Apart from images generated by GANs, a lot of diffusion models are also recently gaining popularity for image synthesis. Studies in literature have reported that the images generated by GAN algorithms and diffusion models have differences in their characteristics \cite{ricker2022towards}. Therefore, in the future, we are considering extending this work to analyze the fairness of forensic models that can also detect images generated by the diffusion models. Also, there is a large scope for mitigation of these biases from the models to develop fair forensic systems that one can trust when deployed in the real world.

\ifCLASSOPTIONcaptionsoff
  \newpage
\fi

%\IEEEtriggeratref{60}
%\IEEEtriggercmd{\enlargethispage{-16cm}}


\bibliographystyle{IEEEtran}
\bibliography{IEEEtran/references_3_bias}
\end{document}